# Optimizing CNNs for identifying Invasive Pollinator *Apis Mellifera* and finding a ligand drug to Protect California's Biodiversity

Author: Arnav Swaroop[1]

Abstract

In North America, there are many diverse species of native bees crucial for the environment, who are the primary pollinators of most native floral species. The Californian agriculture industry imports European honeybees (*Apis Mellifera*) primarily for pollinating almonds. Unfortunately, this has resulted in the unintended consequence of disrupting the native ecosystem and threatening many native bee species as they are outcompeted for food by *A. Mellifera*'s larger colonies. Our first step for protecting the native species is identification with the use of a Convolutional Neural Network (CNN) to differentiate common native bee species from invasive ones. Removing invasive colonies efficiently without harming native species is difficult as pesticides cause myriad diseases in native species. Our approach seeks to prevent the formation of new queens, causing the colony's collapse. Workers secrete royal jelly, a substance that causes fertility and longevity; it is fed to future honeybee queens. Targeting the production of this substance is safe as no native species use it; small organic molecules (ligands) prevent the proteins Apisimin and MRJP1 from combining and producing an oligomer used to form the substance. Ideal ligands bind to only one of these proteins preventing them from joining together: they have a high affinity for one receptor and a significantly lower affinity for the other. We optimized the CNN to provide a framework for creating Machine Learning models that excel at differentiating between subspecies of insects by measuring the effects of image alteration and class grouping on model performance. The CNN is able to achieve an accuracy of 82% in differentiating between invasive and native bee species; 3 ligands have been identified as effective. Our new approach offers a promising solution to curb the spread of invasive bees within California through an identification and neutralization method.

[1] The Harker School, San Jose, CA, USA

## Summary

In this two-part project, we used computational methods to create a solution for the invasive bee problem in California. First, we used machine learning to create an identification app. During this process, we also analyzed how models should be tweaked for optimum performance in differentiation between visually similar subspecies. Next, we use computational biology to identify possible drugs to curb the growth of the invasive species. This two-step method of identification and queen control has potential to be applied to a variety of colony structured invasive insect species.

## Introduction

The decline of native bees

In recent times, conservation news has shed light on the rapidly decreasing populations of honeybees throughout the world due to climate change (CCD: Colony Collapse Disorder) and the threat their extinction could pose for humans due to the fact that over 80% of terrestrial plant species require an animal pollinator (like bees) [1]. An issue that is not nearly as publicized is the even more rapid decline and fragility of native bee populations, which pollinate many of the plants in the United States and provide 3 billion dollars in fruit through pollination [2] and 9 billion dollars in pollination overall [3] per year. Nearly half of native bees specialize in moving pollen around and rely on it as a food source, helping pollinate specialty crops like squash [4]. California is home to nearly half of the United States' native bee species, most of them being endemic. One of the greatest threats to their survival is the foreign introduction of European Honeybees into the ecosystem. Many farmers import or rent honeybees (mostly the european subspecies) to pollinate almond trees [1]. These bees frequently compete with native bees for food and can outcompete them due to their larger colony size and greater aggression. Other uses of honeybees include their namesake honey production which leads to apiaries in agricultural areas. Invasive bee species have been shown to alter the ecosystems they invade and create environmental instability. They disrupt the mutualistic relations between existing fauna and flora, as seen by a study which found that while the introduction of foreign honeybees aids pollination in several plant species, the plants closest to the apiaries die after even a short period of time as a result in pollinating preference and disturbance to the ecosystem [1]. Another issue is that they are adapted to pollinating different plants and have preferences different than native bees, thereby causing change to the landscape of dominant flora [1]. Recently, the rusty patched native bee, the bumblebee (*bombus affinis*) was classified as endangered in California. This signals the imminent danger of losing major native bees permanently, which could dramatically reduce agricultural efficiency as bumblebees pollinate around 30% of crops [5]. Aside from Bumblebees, most native bees in California are solitary.

Our Proposed Solution

To be able to effectively protect the native biodiversity of California while minimizing negative effects on agriculture, invasive bees must be able to be detected easily and naturally controlled within the ecosystem without causing damage to the already fragile native pollinator populations, thereby minimizing damage to apiaries used for specialty pollination like almond trees. We aim to accomplish this by first creating a simple and efficient method of identification and then deploying a localized small molecular drug (ligand) to stop the spread of the invasive species.

The identification is accomplished by a mobile app with a Convolutional Neural Network (CNN) to maximize accuracy while minimizing cost and time. For identification, convolutional neural networks (CNNs) are ideal due to their classification speed. Artificial neural networks simulate biological neural networks with machine learning algorithms. They are comprised of input, hidden (where computation happens), and output layers, each of which contains several nodes. The nodes behave like individual neurons. Nodes in each layer are given weights, which determine the importance of the input based on the strength of the connection, and thresholds, which set a minimum input required to activate the node. If an input signal above the threshold is entered into the neural network, data passes from the first node to a node in the layer above. CNNs consist of convolutional layers, pooling layers, and fully connected layers. Convolutional layers group pixels in the image using weighted filters and try to identify features. These layers help the model quickly analyze larger numbers of pixels, thereby resulting in rapid analysis of details. Pooling layers use unweighted filters to downsize data input. Fully connected layers are the final layers of the CNN and do the classification [6].

The best target for the molecular drug is a protein called MRJP1 which is responsible for creating an important substance called royal jelly which is fed to future queens to prolong their lifespan and make them fertile. For control, MRJP1 forms a complex with another protein, Apisimin so the ligand should aim to bind well to the MRJP1 receptor, preventing the oligomer from forming. Royal jelly is not used in any native species so this method minimizes harm to native species.

## Procedures and Methods

Experimental Setup

For the machine learning component, we chose to use yoloV2 through CoreML and its integrations as this allows for the simple creation of CNNs that can be easily implemented into IOS apps for use on Apple's mobile devices. It was ideal for generating the image classification model and it simplified the data input process as the data is automatically split.

For docking simulations, we used Autodock Vina [7]. Autodock is an open-source protein docking suite. Vina is an addon that allows for more accurate docking and faster simulation. We used this software to prepare the proteins for simulation, to run the simulation itself, and to analyze the results. We used PyMol [8] for visualizing and isolating receptors from the oligomeric structure found on the PDB. We used OpenBabel [9] to convert files of the prepared ligands so that they could be accessed by Vina. For finding ligand and receptor structure, we used the RCSB PDB [10], a protein database containing many protein structures generated using X-ray crystallography.

Use of Convolutional Neural Network

Using CNNs to classify bee species is quite difficult due to the lack of databases with images of bees to train the models and the visual similarity between invasive and native species; the similar color and body shape confuses models.

To reduce the chances of other factors changing the CNN's accuracy, we used yoloV2 and deployed the network into a classifier app using Apple's CoreML tool suite. For the source of the data, we downloaded ~4000 images total of *A. mellifera* and the five groups of native bee species (the native bees were grouped by family). We manually removed anomalous images (clipart, wrong species, etc.) because these would lower the accuracy of the CNN. After removing these images, we were left with 850 images of native bees and 295 images of invasive bees. Using the process outlined in Figure 1 we created machine learning models that were tested for their accuracy, precision, and recall on a random test set.

For each model, we modified the number of classes the families were grouped into and used image alteration for model testing. The 6-class model trained on unaltered images served as the control for the experiment. The CNN was assessed using accuracy, precision, and recall on a test set entirely distinct from the training set which contained the same images but grouped into different classes as per model. Accuracy is the percentage of the test set classified correctly; Precision is defined as True Positives / (True Positives + False Positives); and Recall is defined as True Positives / (True

Positives + False Negatives). For precision and recall, we used the precision and recall of the invasive species to measure the effectiveness of distinguishing them from native bees. It is important for the model to have a higher recall of the invasive species than the native bees as due to the specificity of the proposed drug, native species would be unaffected by it; the safer option is to apply the drug treatment and the model should reflect this.

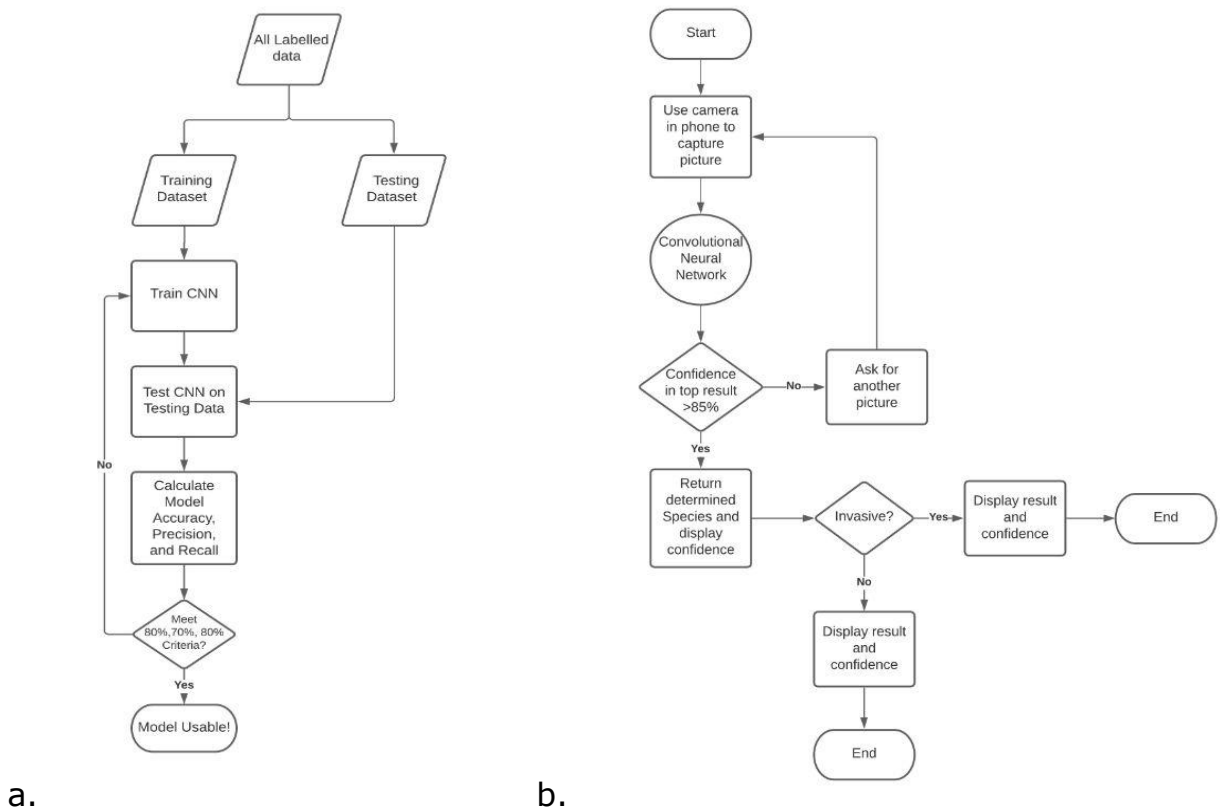

a.  b.

Figure 1: a. flowchart of the process for creating the CNN to meet the criteria, b. flowchart of basic functionality of the app.

Docking Simulations

First, we use PyMOL visualizing software to isolate MRJP1 and Apisimin receptors from the MRJP1-Apisimin oligomer [11] taken from the PDB file by deleting the other chains and pre-existing ligands (isolated receptors shown in Fig. 2). This ensures that future steps of preparing the individual proteins for docking can be done properly. Next, we open these files in the Autodock Vina simulation software and prepare the receptors for docking tests. This consists of adding charges, detecting and repairing missing atoms, adding atom type, and converting the receptor file into the Autodock Vina format.

We then generate grid parameters for receptors to control which part of the protein is considered the receptor that the ligand tries to bind to.

Next, we prepare the ligands for docking. First, like with the receptors, we add charges that may be missing. Next, we calculate and add torsional properties to the molecule to be able to include more binding poses. To accelerate the simulation process, we create a configuration file using the grid parameters and use a script to run the simulations and collect the data.

Last, we evaluate using Bonding Affinity and RMSD values. The docking simulation's results are interpreted by first picking an RMSD upper bound (root mean square distance from the original position) value within which modes are considered. In other words, RMSD represents an acceptable amount of change to the receptor from the bonding with the ligand. The remaining data is compiled and analyzed as we search for a ligand with a substantial difference between its binding affinity to MRJP1 and Apisimin. This is to ensure it can properly disrupt the creation of the MRJP1 complex by making one protein unable to bind to the other.

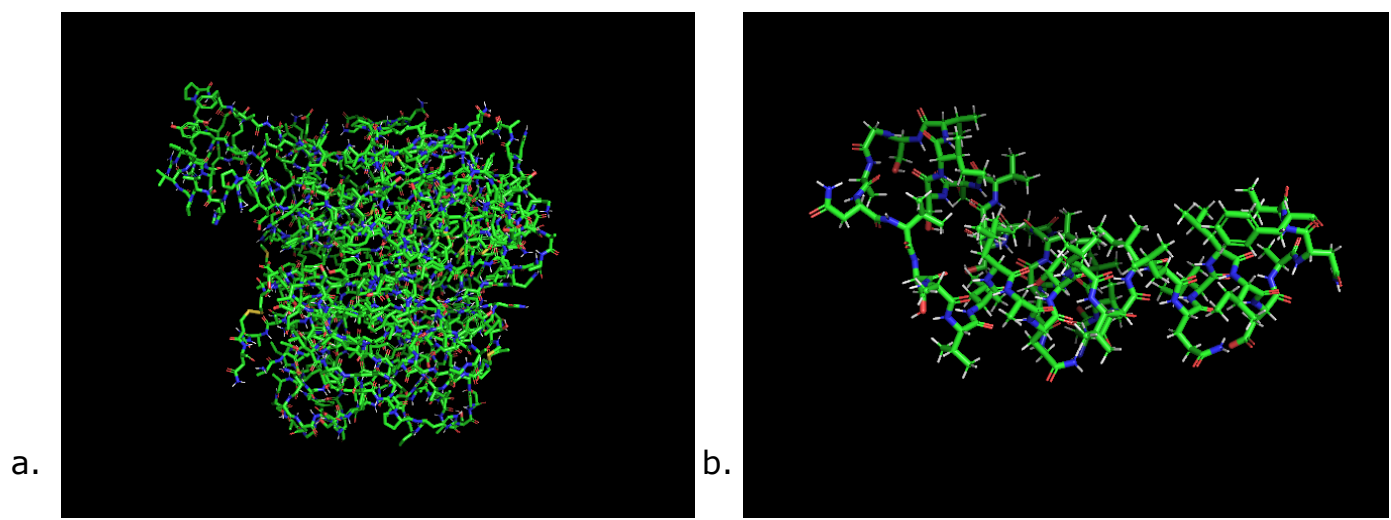

a.　　　　　　　　　　　　　　　　　　　b.

Figure 2: a. Isolated MRJP1 [11], b. Isolated Apisimin [11] (Visualizer: PyMOL [8]).

## Results and Discussion

Our Best Model

The original target criteria for the accuracy was 80% as this value would ensure fast detection trustworthy enough to prompt further investigation but without overfitting to internet images instead of real-time images which would most likely be taken with smartphone cameras rather than professional cameras. As can be seen in Figure 3, the model was able to score an 82% on the final testing set. This version of the model seems to meet those criteria, implying its usability. The precision of the native bee class being significantly higher than that of the invasive species class shows that the CNN is better at minimizing false positives for native bee species than for invasive species. The data also shows that the CNN excels at minimizing false negatives in the invasive species (see Figure 3). For the project, this means that the model selected is best for detecting invasive species while it may not be able to distinguish and identify native species as readily.

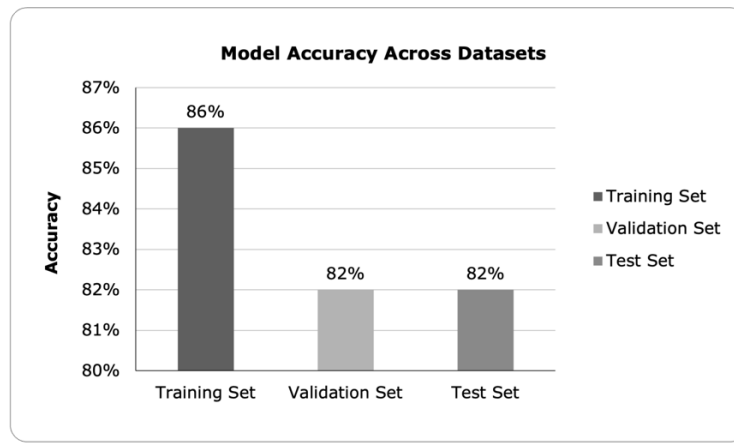

Figure 2: The accuracy data of the machine learning model per dataset.

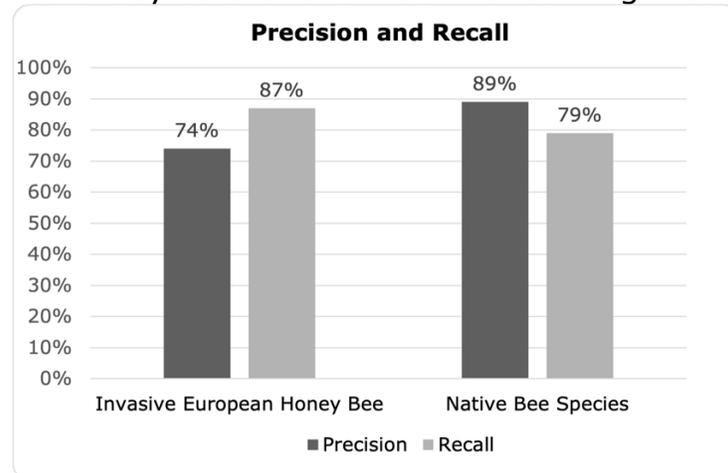

Figure 3: Precision and recall for the 2 classes of the model.

## Comparison of CNN Models

In most cases, CNN model performance is inversely related to the number of classes present. As such, we expected to find that the 2-class model would have the greatest performance, followed by the 3-class model, and finally the 6-class model. Interestingly, we found that the 3-class model where native bee species were split into two groups based on genetic and physiological similarity (see Figure 4) performed the best overall. This is likely due to the incredible diversity of native bees in the state. The similarities in the shapes of mason and carpenter bees, which have a rounder, stouter appearance, helps the model recognize them distinctly. In addition, leafcutter bee, bumblebee, and sweat bee species often have distinctive lines of white fuzzy hairs on their abdomen, and several have the light-colored fuzz on their thorax as well. These physiological differences help separate these groups of bees from the thinner A. mellifera which have lustrous abdomens and yellow fuzz only on their thorax. For the 2-class model, the native class included many diverse species of native bees and thus, the model struggled to recognize them as seen by the low recall of around 60% The 2-class model's precision for the invasive species class is incredibly high because it was able to easily recognize the invasive species which had a constant appearance.

    Image alteration is used to increase the effectiveness of CNNs by preventing overfitting, a condition that occurs when the validation set includes training set images or images too similar to the training set. However, since bees always fly upright and in a certain posture, we predicted that alteration would not be helpful as the sets were already diverse in terms of orientation of the bees. As seen in Figure 5, image alteration had negligible effects on the precision and accuracy of the model, showing that it did not impact the rates of false positives or the accuracy of the model. However, using image alteration did significantly improve recall corresponding to an increase in recognition of the invasive species. This is likely due to a bias in the training set of position and background.

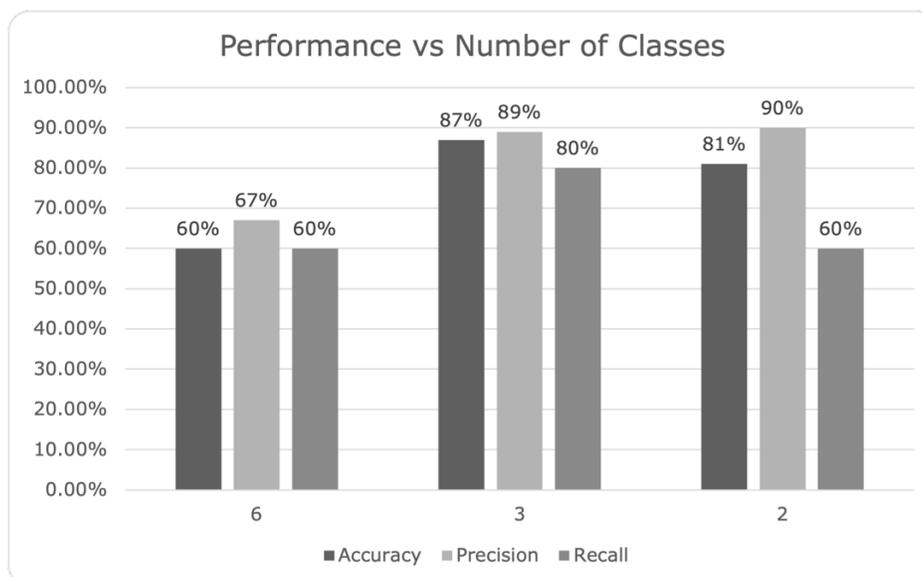

Figure 4: Effects of Number of Classes on Performance

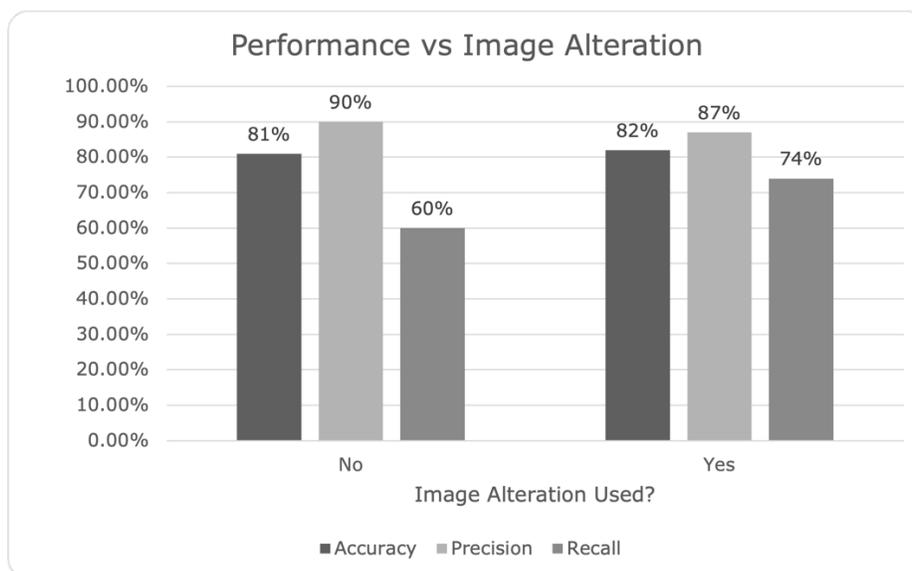

Figure 5: Effects of Image Manipulation on Performance

Docking Simulations

The data for the docking simulations was more complex to analyze; each of the simulations returned a table containing 9 poses of binding, returning the affinity of the bond (negative is better), and the RMSD value (the ligand's distance from the most optimal bonding position in Angstroms) per ligand. For this simulation, in order to preserve a high standard of data, the criteria for the data to be used in analysis was an RMSD upper bound value of under 3.5 Å. This constraint was put in place because above that value, the binding simulations are inaccurate and further difference in RMSD (i.e., between RMSD values of 4 and 8) changes little. After this selection, the data was condensed into a table (please see Table 1 in the Appendix) and then the averages of the binding affinities of each ligand were taken and graphed as shown in Fig. 6 to simplify qualitative analysis.

## Conclusions

We found that for optimizing a CNN for the purpose of recognizing an invasive subspecies of insect, the greatest consideration is the diversity of native species. If the diversity is large, the CNN will achieve best performance if the native species are split into 2 or 3 subgroups based on appearance and differentiating characteristics. Otherwise, creating a 2-class invasive vs. native model is ideal. As for image alteration, it is recommended even though it does not cause significant change to accuracy or precision because it does significantly boost recall for the invasive species.

The docking simulation results show that the best candidate ligands are VD3 and D2V because while their binding affinity for Apisimin is around that of the control, their binding affinity for MRJP1 is high compared to the control. CLR and ERG are also possible options and should not be ruled out without a wet lab trial. A high binding affinity (more negative value) for one protein makes it harder for another protein to also bind to the same ligand because a stronger bond between the ligand and the receptor leaves less of the surface area of the ligand able to interact with the receptor. The most effective method of deploying these ligands would be to put them in a solution (with pH and salt levels similar to that of the invasive bees' hemolymph) and spray it on all flowers in the swath of area where the invasive bees have been detected. This would ensure that the workers would take back some solution containing our drug to their hive and spread it around causing all the workers to be unable to produce the Royal Jelly.

## Future Work

A new and emerging method of small molecular drug development is using molecular *de novo* design to use machine learning to create the ligand from the receptor. Unfortunately, this new molecule could potentially create issues for native species so rigorous testing is required.

For the app, the first invasive vs native CNN could potentially feed into another CNN to determine which species it is and display some interesting facts about it. After gaining a large user base, the app could also help active conservation surveys by tagging native, endangered species in a database after being detected by a user. These modifications would help the system work better, helping native bees, biodiversity as a whole, and raising awareness about current ecological issues.

# Appendix
Table 1: Table of all data that passed the RMSD criteria.

| MRJP1 Ligand | Mode | Binding Affinity (kcal/mol) | RSMD UB (Å) | Apisimin Ligand | Mode | Binding Affinity (kcal/mol) | RSMD UB (Å) |
|---|---|---|---|---|---|---|---|
| 94R (Control) | 1 | -6.0 | 0.000 | 94R (Control) | 1 | -4.7 | 0.000 |
| 94R (Control) | 2 | -5.8 | 2.958 | CLR | 1 | -5.3 | 0.000 |
| 94R (Control) | 7 | -5.2 | 2.250 | VD3 | 1 | -5.2 | 0.000 |
| CLR | 1 | -6.7 | 0.000 | MHQ | 1 | -5.4 | 0.000 |
| CLR | 2 | -6.4 | 3.168 | MHQ | 8 | -4.7 | 3.445 |
| CLR | 4 | -6.0 | 2.320 | DVE | 1 | -5.8 | 0.000 |
| CLR | 7 | -5.8 | 2.736 | DVE | 2 | -5.6 | 2.670 |
| VD3 | 1 | -6.9 | 0.000 | ERG | 1 | -5.8 | 0.000 |
| VD3 | 2 | -6.8 | 2.719 | ERG | 4 | -5.3 | 2.067 |
| MHQ | 1 | -6.5 | 0.000 | ERG | 9 | -4.7 | 3.030 |
| MHQ | 3 | -6.2 | 3.413 | D2V | 1 | -5.4 | 0.000 |
| MHQ | 8 | -5.6 | 3.319 | D2V | 4 | -4.6 | 3.168 |
| DVE | 1 | -7.1 | 0.000 | LNP | 1 | -5.4 | 0.000 |
| DVE | 2 | -6.3 | 2.910 | LAN | 1 | -5.9 | 0.000 |
| DVE | 7 | -5.6 | 2.813 | LAN | 2 | -5.7 | 2.275 |
| ERG | 1 | -6.9 | 0.000 | | | | |
| ERG | 3 | -6.2 | 2.405 | | | | |
| ERG | 4 | -6.0 | 2.919 | | | | |
| D2V | 1 | -6.9 | 0.000 | | | | |
| D2V | 4 | -6.3 | 3.401 | | | | |
| LNP | 1 | -6.5 | 0.000 | | | | |
| LNP | 2 | -6.1 | 2.249 | | | | |
| LNP | 4 | -5.6 | 2.618 | | | | |
| LNP | 5 | -5.5 | 2.457 | | | | |
| LNP | 7 | -5.3 | 3.245 | | | | |
| LAN | 1 | -7.1 | 0.000 | | | | |
| LAN | 5 | -6.4 | 3.236 | | | | |
| LAN | 9 | -6.2 | 2.746 | | | | |

## Bibliography

1. McAfee, A. (2020, November 4). The Problem with Honey Bees. Scientific American. https://www.scientificamerican.com/article/the-problem-with-honey-bees/
2. Kopec, K., & Burd, L. A. (2017, February). Pollinators in Peril: A systematic status review of North American and Hawaiian native bees. Center for Biological Diversity. https://www.biologicaldiversity.org/campaigns/native_pollinators/pdfs/Pollinators_in_Peril.pdf
3. Protecting California's Native Bees. (2018, October 23). Medium. Retrieved October 13, 2021, from https://medium.com/wild-without-end/protecting-californias-native-bees-4308268ce438
4. What is the role of native bees in the United States? (n.d.). U.S. Geological Survey. Retrieved October 13, 2021, from https://www.usgs.gov/faqs/what-role-native-bees-united-states?qt-news_science_products=0#qt-news_science_products
5. Native bees often better pollinators than honey bee. Berkeley Research. https://vcresearch.berkeley.edu/news/native-bees-often-better-pollinators-honey-bee
6. IBM Cloud Education. (n.d.). What are convolutional neural networks? IBM. Retrieved September 21, 2022, from https://www.ibm.com/cloud/learn/convolutional-neural-networks
7. O. Trott, A. J. Olson, AutoDock Vina: improving the speed and accuracy of docking with a new scoring function, efficient optimization and multithreading, Journal of Computational Chemistry 31 (2010) 455-461
8. The PyMOL Molecular Graphics System, Version 2.0 Schrödinger, LLC. https://pymol.org (accessed Dec 2021)
9. N M O'Boyle, M Banck, C A James, C Morley, T Vandermeersch, and G R Hutchison. "Open Babel: An open chemical toolbox." J. Cheminf. (2011), 3, 33. DOI:10.1186/1758-2946-3-33
10. H.M. Berman, J. Westbrook, Z. Feng, G. Gilliland, T.N. Bhat, H. Weissig, I.N. Shindyalov, P.E. Bourne.
11. (2000) The Protein Data Bank Nucleic Acids Research, 28: 235-242.
12. Tian, W., Li, M., Guo, H., Peng, W., Xue, X., Hu, Y., Liu, Y., Zhao, Y., Fang, X., Wang, K., Li, X., Tong, Y., Conlon, M. A., Wu, W., Ren, F., & Chen, Z. (2018). Architecture of the native major royal jelly protein 1 oligomer. Nature Communications, 9(1). https://doi.org/10.1038/s41467-018-05619-1
13. Underwood, E. (2021, June 24). Small Wonders: The Plight and Promise of California's Native Bees. Flora Magazine. Retrieved October

13, 2021, from https://www.cnps.org/flora-magazine/small-wonders-the-plight-and-promise-of-californias-native-bees-23883.
14. Koch, J., Lozier, J., Strange, J., Ikerd, H., Griswold, T., Cordes, N., Solter, L., Stewart, I., & Cameron, S. (2015). USBombus, a database of contemporary survey data for north american bumble bees (Hymenoptera, apidae, bombus) distributed in the united states. Biodiversity Data Journal, 3, e6833. https://doi.org/10.3897/BDJ.3.e6833
15. Gupta, D. (2020, January 30). Fundamentals of Deep Learning – Activation Functions and When to Use Them? Analytics Vidhya. Retrieved October 13, 2021, from https://www.analyticsvidhya.com/blog/2020/01/fundamentals-deep-learning-activation-functions-when-to-use-them/
16. How Much Does It Cost For Bee Removal? (n.d.). HomeAdvisor. Retrieved October 13, 2021, from https://www.homeadvisor.com/cost/environmental-safety/bee-removal/#:~:text=%240-,Bee%20and%20Beehive%20Removal%20Prices,or%20more%20than%20an%20exterminator.
17. Gorman, C. E., Potts, B. M., Schweitzer, J. A., & Bailey, J. K. (2014). Shifts in species interactions due to the evolution of functional differences between endemics and non-endemics: An endemic syndrome hypothesis. PLoS ONE, 9(10), e111190. https://doi.org/10.1371/journal.pone.0111190
18. About Honey Bees - Types, Races, and Anatomy. (n.d.). University of Arkansas System: Division of Agriculture: Research & Extension. https://www.uaex.edu/ farm-ranch/special-programs/beekeeping/about-honey-bees.aspx Vance, E. (n.d.).